\documentclass{article}

\PassOptionsToPackage{numbers, compress}{natbib}
\usepackage[preprint]{neurips_2026}

\usepackage[utf8]{inputenc}
\usepackage[T1]{fontenc}
\usepackage{hyperref}
\usepackage{url}
\usepackage{booktabs}
\usepackage{amsfonts}
\usepackage{amsmath}
\usepackage{nicefrac}
\usepackage{microtype}
\usepackage{xcolor}
\usepackage{graphicx}

\title{Representation over Routing: Diagnosing Temporal Routing Pathologies in Multi-Timescale PPO}

\author{%
  Jing Sun \\
  Information Engineering School\\
  Chengyi College, Jimei University\\
  Xiamen 361000 \\
  \texttt{dlwlrma@jmu.edu.cn, ben.dlwlrma@gmail.com} \\
}

\begin{document}

\maketitle

\begin{abstract}
Temporal credit assignment in reinforcement learning is often approached by introducing value estimates at multiple discount factors. A natural next step is to let the actor dynamically route among these temporal heads, using either differentiable attention or heuristic uncertainty weights. This paper argues that such routing can create a numerical shortcut rather than a reliable temporal abstraction. We study this issue in a controlled PPO setting on LunarLander-v2, using the environment as a visual sandbox for diagnosing failure modes. First, we formalize Surrogate Objective Hacking: a differentiable softmax router exposed to the PPO surrogate receives a direct gradient toward advantage heads that are numerically favorable for the current update, even when this routing change does not correspond to improved physical control. Because unnormalized advantages at different discount factors have different effective scales, this creates a scale-discrepancy vulnerability. Second, we identify the Paradox of Temporal Uncertainty in gradient-free error-based routing: short-horizon heads can receive the largest routing share because their prediction targets are easier, even when they are less aligned with delayed task success. As a structural response, we study Target Decoupling: the critic may retain multi-timescale auxiliary heads, but the actor is updated only with the long-horizon advantage. Target Decoupling is not presented as a broad performance booster; in this run set it removes the exploitable actor-side routing pathway and improves the observed worst-seed return. Code is available at \url{https://github.com/ben-dlwlrma/Representation-Over-Routing}.
\end{abstract}

\section{Introduction}
\label{Introduction}

Temporal credit assignment remains a central difficulty in reinforcement learning (RL). A single discount factor $\gamma$ imposes one temporal scale on both value prediction and policy improvement. Short horizons provide dense and low-variance learning signals, but can bias the policy toward immediate reward shaping. Long horizons better match delayed objectives, but their targets are harder to estimate early in training. Multi-timescale value prediction is therefore appealing: a critic can represent several discounted futures in parallel, for example
\begin{equation}
\Gamma=\{0.5,0.9,0.99,0.999\}.
\end{equation}

The main question is not whether these heads can be predicted. The more delicate question is where they should enter the actor update. A common design instinct is to route among the heads dynamically, allowing the actor to combine short-term and long-term advantages using state-dependent weights. This paper studies a failure mode of that design. When the router is differentiable and participates in the PPO surrogate objective, the actor can improve the proxy objective by changing the routing weights themselves. This optimization channel need not correspond to a stable physical abstraction of temporal context.

We call this a scale-discrepancy vulnerability. Advantages estimated at different discount factors are not intrinsically calibrated. Their effective horizons, target magnitudes, and variances differ. A softmax router exposed to these unnormalized signals can therefore learn to select whichever head is numerically favorable for the surrogate update, rather than learning a semantically meaningful horizon selector. We refer to this failure mode as Surrogate Objective Hacking.

We use LunarLander-v2 as a controlled diagnostic environment. It is not intended here as a broad benchmark claim. The environment is useful because it exposes a clear tension between immediate shaping terms and delayed landing outcomes, and its failure modes are visually interpretable. Our empirical study has three parts. First, we diagnose differentiable actor-side routing using a three-panel diagnostic: episodic return, HackRate, and attention entropy. The learned router initially tracks the maximum-advantage head at a rate above the random baseline, then quickly collapses to near-zero entropy, while the policy fails to solve the task. Second, we examine a gradient-free error-based router and find evidence of a different observed degeneration: low-error short-horizon predictions can dominate the weighting rule and bias behavior toward myopic control. We refer to this second failure mode as the Paradox of Temporal Uncertainty. Third, we evaluate Target Decoupling, a structural separation principle in which the actor ignores routing and uses only the long-horizon advantage, while the critic retains auxiliary short-horizon heads as regularizers.

The contribution of this paper is therefore diagnostic rather than performance-driven:
\begin{itemize}
    \item We formalize Surrogate Objective Hacking through the router-gradient mechanism by which differentiable temporal routing can be exploited inside a PPO surrogate objective.
    \item We identify the Paradox of Temporal Uncertainty in error-based routing, where prediction reliability can become misaligned with temporal relevance.
    \item We evaluate Target Decoupling as a structural separation principle. In this run set it improves the observed worst-seed return and reduces observed across-seed dispersion, while the original hypothesis that auxiliary heads reduce within-batch long-advantage variance is not supported by the data.
\end{itemize}

\section{Related Work}
\label{related_work}

\textbf{Temporal credit assignment and multi-horizon value prediction.}
Generalized Advantage Estimation (GAE) \cite{schulmanHighDimensionalContinuousControl2018} reduces variance by combining temporal-difference residuals with an exponentially decaying trace, but it still depends on a chosen discount factor. Multi-horizon value representations are motivated by both algorithmic considerations and neurobiological observations of distributed temporal coding in dopamine systems \cite{dabneyDistributionalCodeValue2020}. Our work is aligned with this motivation, but focuses on a specific question: whether multi-horizon predictions should be routed into the actor objective.

\textbf{Routing, attention, and proxy optimization.}
Attention and gating mechanisms are widely used to adapt computation to state. In policy-gradient methods, however, a differentiable router inside the actor objective is itself optimized by the policy loss. This creates a possible proxy-optimization channel: the router can alter the scalar advantage used by the surrogate without improving physical control. This failure mode is related in spirit to reward hacking and specification gaming \cite{amodeiConcreteProblemsAI2016}, but occurs inside the temporal aggregation mechanism rather than in the external reward.

\textbf{Uncertainty weighting.}
Uncertainty and ensemble weighting are often used to reduce the influence of unreliable estimates, as in methods such as SUNRISE \cite{leeSUNRISESimpleUnified2021}. In a multi-timescale setting, however, prediction error is coupled to horizon length. Short-horizon heads can have smaller errors because their targets are easier, not because they are better aligned with the long-term objective. This creates a route to myopic degeneration when error is used directly as a routing criterion.

\section{Methodology}
\label{methodology}

We first describe the multi-timescale critic, then derive the scale-mismatch mechanism behind differentiable routing, and finally define Target Decoupling.

\subsection{Multi-Timescale Value Representation}

Let the critic output a vector of value predictions,
\begin{equation}
V_\psi(s_t)=
\left[
V_{\gamma_1}(s_t),
V_{\gamma_2}(s_t),
\ldots,
V_{\gamma_K}(s_t)
\right].
\end{equation}
For each discount factor $\gamma_i$, we compute a GAE estimate $\hat{A}_{\gamma_i}(s_t,a_t)$ and a value target $\hat{R}_{\gamma_i}(s_t)$. The standard multi-head critic objective averages losses across heads:
\begin{equation}
\mathcal{L}_{V}(\psi)
=
\frac{1}{K}
\sum_{i=1}^{K}
\frac{1}{2}
\left(
V_{\gamma_i}(s_t)-\hat{R}_{\gamma_i}(s_t)
\right)^2.
\end{equation}
This objective can be useful as auxiliary representation learning. The issue studied here arises when the actor also learns to route among the heads.

\subsection{Router-Gradient Mechanism}

Consider a differentiable router with weights
\begin{equation}
w_i(s_t;\phi)
=
\frac{\exp(z_i(s_t;\phi))}
{\sum_{j=1}^{K}\exp(z_j(s_t;\phi))},
\end{equation}
where $z_i$ are router logits. The routed advantage used by the actor is
\begin{equation}
\hat{A}_{w}(s_t,a_t)
=
\sum_{i=1}^{K}
w_i(s_t;\phi)\hat{A}_{\gamma_i}(s_t,a_t).
\end{equation}
For a fixed sample, the derivative of the routed advantage with respect to a router logit is
\begin{equation}
\frac{\partial \hat{A}_{w}}{\partial z_j}
=
w_j
\left(
\hat{A}_{\gamma_j}-\hat{A}_{w}
\right).
\label{eq:router_gradient}
\end{equation}

Equation \ref{eq:router_gradient} shows that the router receives a direct gradient toward heads whose signed advantage is larger than the current routed mixture. In the unclipped region of PPO, the surrogate contains the ratio-weighted advantage term,
\begin{equation}
\mathcal{L}_{\mathrm{clip}}
\approx
\mathbb{E}_{t}
\left[
r_t(\theta)\hat{A}_{w}(s_t,a_t)
\right],
\end{equation}
so the router parameters are updated through
\begin{equation}
\nabla_{\phi}\mathcal{L}_{\mathrm{clip}}
\approx
\mathbb{E}_{t}
\left[
r_t(\theta)
\sum_{i=1}^{K}
\hat{A}_{\gamma_i}(s_t,a_t)
\nabla_{\phi}w_i(s_t;\phi)
\right].
\end{equation}
Thus, actor-side routing is not a passive interpretation layer. It is an optimization variable inside the surrogate objective.

\subsection{Cross-Horizon Scale Mismatch}

Cross-horizon advantages are not automatically on a common numerical scale. If rewards are bounded as $|r_t|\leq R_{\max}$, then discounted values satisfy
\begin{equation}
|V^\gamma(s)|
\leq
\frac{R_{\max}}{1-\gamma}.
\label{eq:value_bound}
\end{equation}
For a finite effective horizon $H$, the corresponding bound is
\begin{equation}
|V^\gamma_H(s)|
\leq
R_{\max}
\frac{1-\gamma^H}{1-\gamma}.
\end{equation}
These bounds do not imply that a long-horizon advantage is larger on every sample. They do imply that different discount factors induce different effective scales and variances. GAE estimates inherit this mismatch because their residual traces accumulate over different temporal windows:
\begin{equation}
\hat{A}^{\gamma,\lambda}_t
=
\sum_{\ell=0}^{\infty}
(\gamma\lambda)^\ell
\delta^{\gamma}_{t+\ell}.
\end{equation}
Without head-wise calibration, the router compares quantities that are not intrinsically comparable. In the implementation studied here, the final routed scalar advantage is batch-normalized before entering the PPO policy loss, as in standard PPO practice. This normalization does not calibrate the router's input decision: the softmax weights are still computed over raw, cross-head advantage values before aggregation into a scalar. The routing decision therefore remains exposed to head-scale mismatch, creating a numerical channel through which the actor can improve the surrogate by reallocating weights among heads rather than by improving action selection.

\subsection{Entropy-Collapse Signature}

The entropy of the router is
\begin{equation}
H(w)
=
-
\sum_{i=1}^{K}
w_i\log w_i.
\end{equation}
When one head is repeatedly favored by the routed surrogate gradient, the corresponding logit can grow relative to the others, moving the softmax toward a vertex of the probability simplex. In that regime,
\begin{equation}
H(w)\rightarrow 0.
\end{equation}
This entropy collapse is the predicted signature of the scale mismatch: the router ceases to represent a graded temporal mixture and becomes a hard selector.

\subsection{Target Decoupling}

Target Decoupling removes the actor-side routing pathway. The critic may retain auxiliary heads, but the actor update uses only the long-horizon advantage:
\begin{equation}
\hat{A}_{\mathrm{actor}}(s_t,a_t)
=
\hat{A}_{\gamma=0.999}(s_t,a_t).
\label{eq:target_decoupling_actor}
\end{equation}
The policy loss is therefore
\begin{equation}
\mathcal{L}^{\mathrm{PPO}}_{\theta}
=
\mathbb{E}_{t}
\left[
\min
\left(
r_t(\theta)\hat{A}_{\mathrm{actor}},
\mathrm{clip}(r_t(\theta),1-\epsilon,1+\epsilon)
\hat{A}_{\mathrm{actor}}
\right)
\right].
\end{equation}
The critic objective can include auxiliary short-horizon losses:
\begin{equation}
\mathcal{L}_{V}
=
\mathcal{L}_{\gamma_{\mathrm{long}}}
+
\lambda
\sum_{\gamma_k\in\Gamma_{\mathrm{short}}}
\mathcal{L}_{\gamma_k}.
\label{eq:aux_loss}
\end{equation}
This design does not train a better router. It removes the actor-side shortcut from the policy objective by construction.

\section{Experiments and Diagnostic Evidence}
\label{Experiments}

All experiments use PPO with MLP actor-critic networks on LunarLander-v2. We use the environment as a controlled diagnostic sandbox because its reward structure combines short-term shaping terms, control penalties, and a delayed landing outcome. Unless stated otherwise, the discount set is
\begin{equation}
\Gamma=\{0.5,0.9,0.99,0.999\}.
\end{equation}
The goal of these experiments is not to establish broad benchmark superiority. The goal is to test whether the predicted routing failure modes appear in a concrete and reproducible PPO setting.

\subsection{Surrogate Exploitation in Differentiable Attention Routing}

We first evaluate an actor-side attention router whose softmax weights multiply the four GAE heads. To diagnose whether the router is tracking the numerically favorable advantage head, we define
\begin{equation}
\mathrm{HackRate}
=
\mathbb{E}
\left[
\mathbf{1}
\left(
\arg\max_i w_i(s_t)
=
\arg\max_i \hat{A}_{\gamma_i}(s_t,a_t)
\right)
\right].
\end{equation}
With four heads, the random-routing baseline is $1/4$.

\begin{figure}[t]
  \centering
  \includegraphics[width=\linewidth]{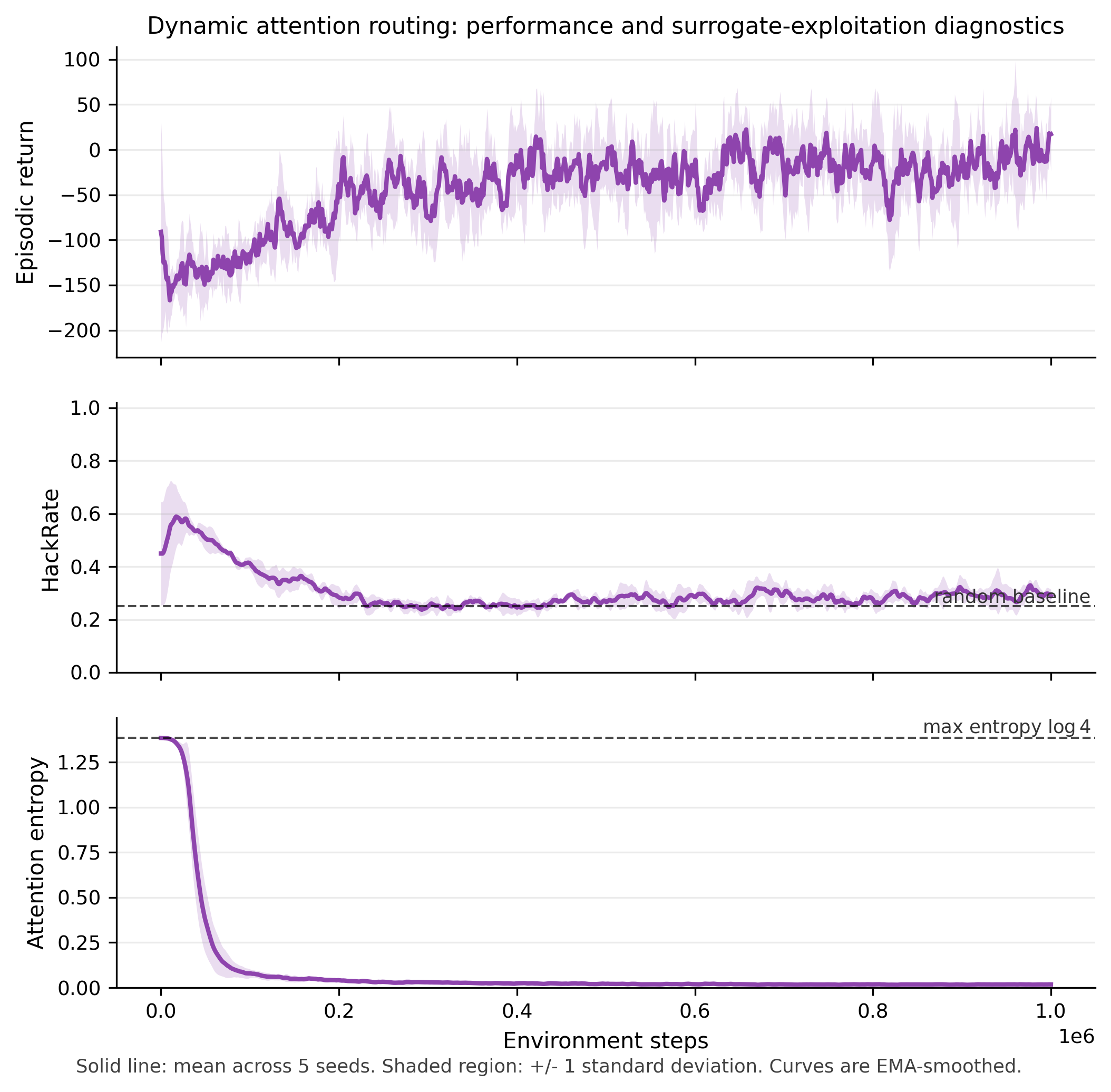}
  \caption{Diagnostic triad for differentiable temporal attention routing over 5 seeds. The top panel shows episodic return. The middle panel shows HackRate, with the dashed line marking the random baseline $1/4$. The bottom panel shows attention entropy, with the dashed line marking the maximum entropy $\log 4$. The router initially selects the maximum-advantage head at a rate above chance, then rapidly collapses to near-zero entropy, while return remains unstable and poor. This pattern is evidence consistent with surrogate exploitation of the routing pathway.}
  \label{fig:surrogate_triad}
\end{figure}

Figure \ref{fig:surrogate_triad} supports the predicted failure mode. Early in training, HackRate rises well above the random baseline, indicating that the router is not behaving as an independent temporal abstraction module. Shortly after, attention entropy collapses toward zero across seeds. The return curve does not show a corresponding improvement in task performance. The evidence is therefore not that high HackRate persists throughout training, but that early surrogate chasing is followed by hard routing collapse and failure to solve the task.

\subsection{The Paradox of Temporal Uncertainty in Error-Based Routing}

A natural response is to remove router gradients and compute weights from prediction errors. We test a gradient-free rule in which heads with smaller absolute temporal-difference error receive larger weights:
\begin{equation}
w_i
=
\frac{\exp(-|\delta_{\gamma_i}|/\tau)}
{\sum_{j=1}^{K}\exp(-|\delta_{\gamma_j}|/\tau)}.
\end{equation}
This avoids direct backpropagation through the router, but introduces a different bias. Short-horizon targets are easier to predict and can have systematically smaller errors. The routing rule can therefore prefer the short-horizon head because it is numerically reliable, not because it is aligned with the long-term landing objective.

\begin{figure}[t]
  \centering
  \includegraphics[width=\linewidth]{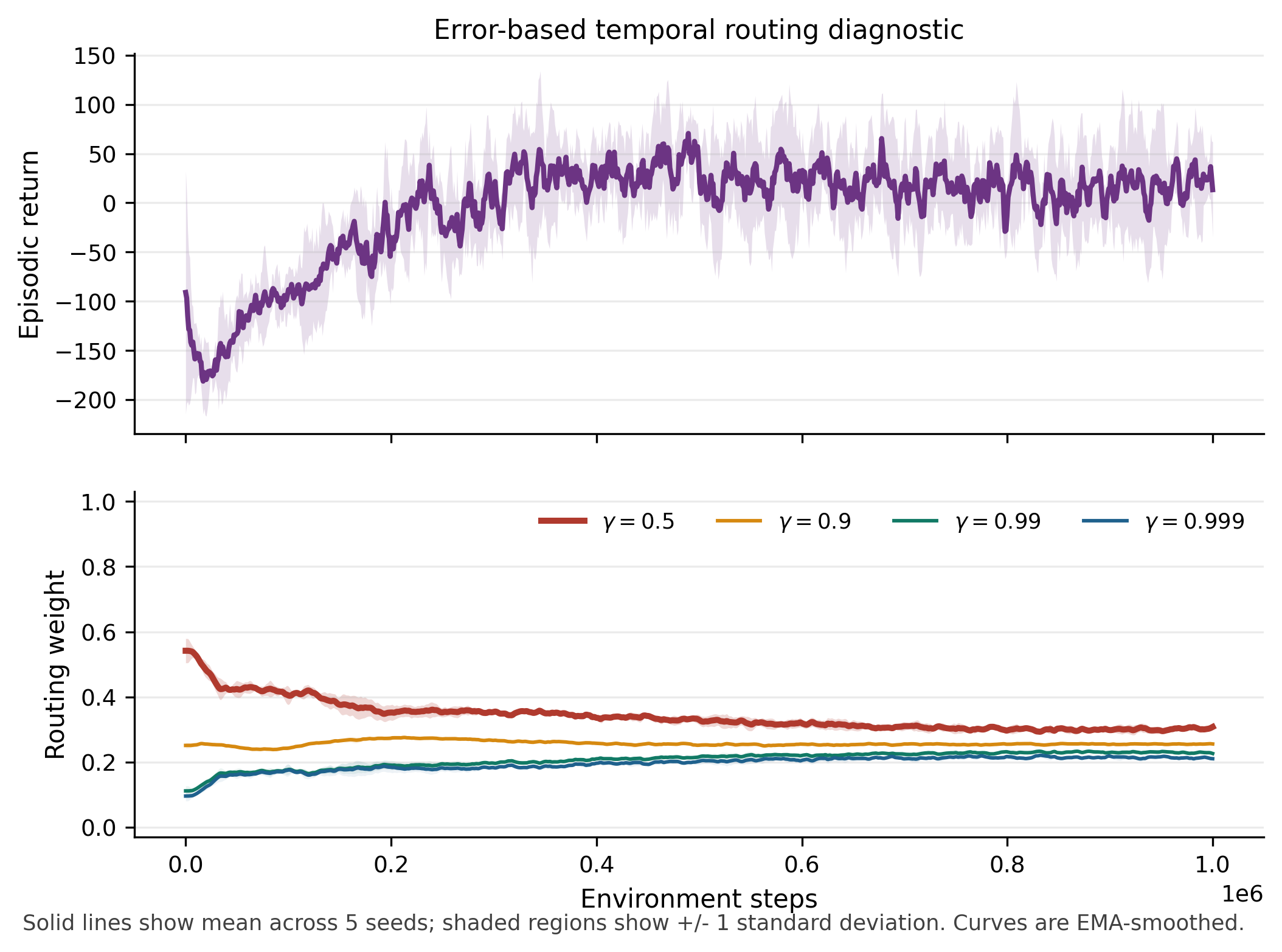}
  \caption{Diagnostic plot for gradient-free error-based routing. Solid lines show mean across 5 seeds, with shaded regions indicating $\pm$ one standard deviation. The top panel shows episodic return, and the bottom panel shows routing weights induced by absolute temporal-difference error. Because short-horizon targets are easier to predict, the $\gamma=0.5$ head can receive dominant weight while return remains poor. The figure supports an observed myopic-degeneration pattern in this PPO setting, not a universal impossibility result for uncertainty-based weighting.}
  \label{fig:error_routing}
\end{figure}

Figure \ref{fig:error_routing} provides the corresponding diagnostic. The short-horizon weight, especially the $\gamma=0.5$ head, receives a large share of the routing mass while episodic return remains poor. This is evidence for an observed myopic-degeneration failure mode in this setup: low-error short-horizon predictions can receive the largest routing share even when they do not support the delayed landing objective. This result should not be read as a claim that all uncertainty weighting fails. It identifies a specific failure mode of using raw prediction error as a temporal routing criterion when horizons have different intrinsic difficulty.

\subsection{Target Decoupling as Structural Separation}

Target Decoupling removes the actor-side temporal router and uses the long-horizon advantage in Equation \ref{eq:target_decoupling_actor}. We also tested the hypothesis that auxiliary heads improve performance by reducing the within-batch variance of the long-horizon advantage. That hypothesis was not supported by the ablation: the measured long-advantage variance curves for auxiliary and non-auxiliary variants became entangled after early training.

The remaining effect is better described as reliability-oriented regularization. Auxiliary heads do not provide a routed policy target. Instead, they constrain the critic through additional prediction tasks while the actor remains aligned with the long-horizon objective.

\begin{figure}[t]
  \centering
  \includegraphics[width=0.82\linewidth]{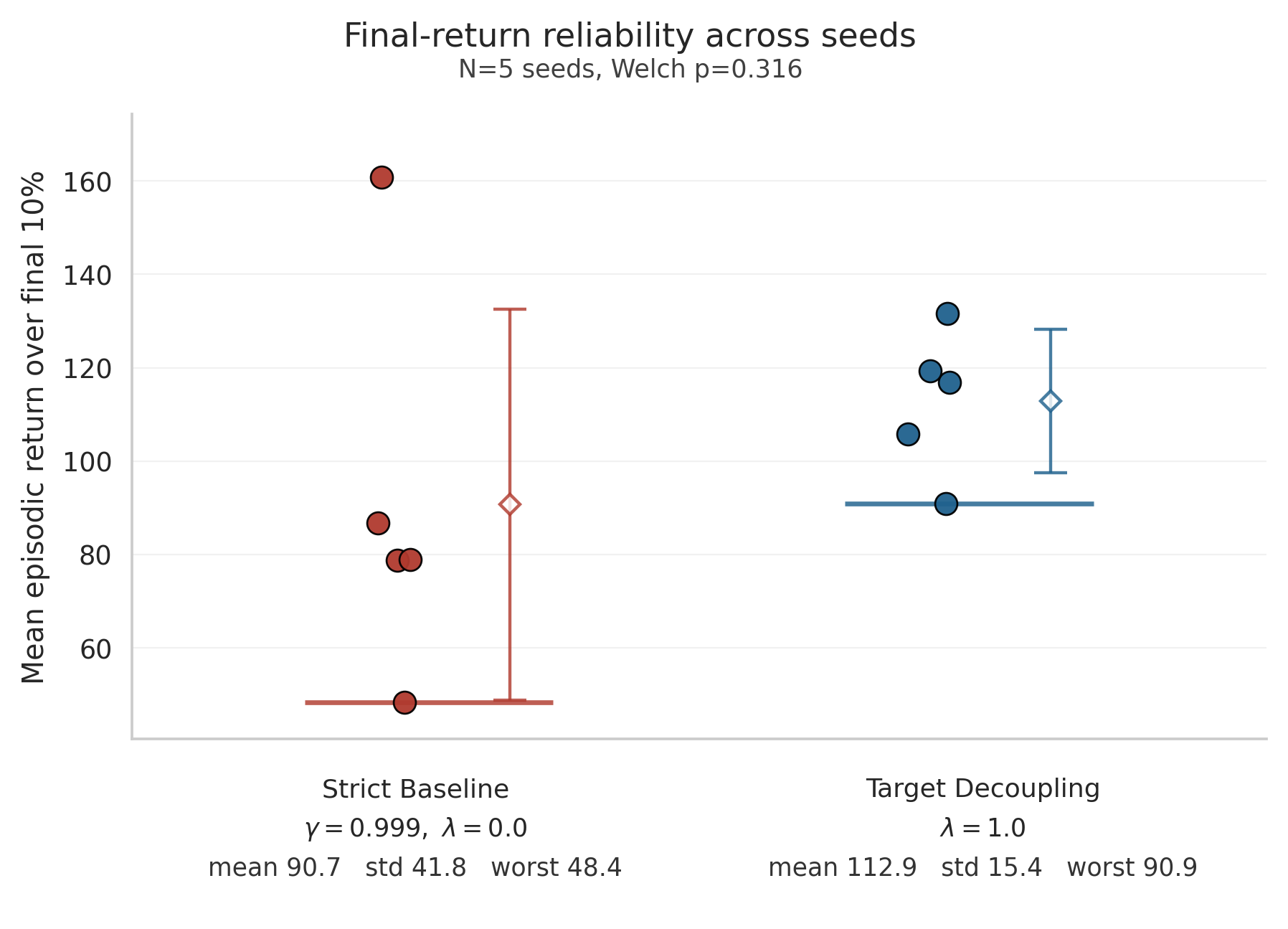}
  \caption{Final-return reliability over 5 seeds, computed from the last 10 percent of training. Each point is one seed. Diamonds and error bars show mean $\pm$ one standard deviation, and the horizontal colored bars mark the worst seed in each condition. The figure is a reliability diagnostic rather than a claim about mean-performance significance. In this controlled setting, Target Decoupling raises the observed worst-seed return and reduces observed across-seed dispersion.}
  \label{fig:reliability_boxplot}
\end{figure}

Figure \ref{fig:reliability_boxplot} shows the reliability effect. The strict long-horizon baseline has a lower worst seed and larger dispersion, whereas Target Decoupling produces a tighter cluster and a higher observed lower bound. This supports the interpretation that structural separation removes an actor-side failure channel and improves the observed worst-seed return in this run set.

\section{Conclusion}

This paper reframes multi-timescale PPO as a question of optimization geometry rather than only temporal representation. Multi-horizon critics can be useful, but actor-side routing over unnormalized advantage heads creates a scale mismatch that can be exploited by the surrogate update. We identify two failure modes that are easy to conflate. Surrogate Objective Hacking is a gradient-path failure: a softmax router inside the PPO surrogate is directly optimized to select numerically favorable heads, and in our controlled LunarLander-v2 study this manifests as early maximum-advantage tracking, rapid entropy collapse, and poor return. The Paradox of Temporal Uncertainty is a gradient-free reliability-versus-relevance failure: error-based routing can prefer low-error short-horizon heads because their targets are easier, even when those heads are less aligned with delayed task success.

Target Decoupling is a structural response: preserve multi-horizon critic prediction as auxiliary regularization, but remove temporal routing from the actor objective. The resulting method is not presented as a general performance booster. Its value in this study is the removal of the identified actor-side routing channel and the improved observed worst-seed return across the tested seeds.

Future work should test whether these pathologies persist beyond PPO and LunarLander-v2, including SAC or TD3 with multi-horizon $Q$ heads, offline RL with extrapolation-biased value estimates, and robotic or safety-critical control tasks where short-horizon reflexes and long-horizon objectives conflict. A second direction is to replace scalar routing with calibrated temporal control mechanisms, such as head-wise advantage calibration, stop-gradient routers, or context-dependent threat appraisal modules that can shift toward short-horizon reflexes under imminent risk without exposing the actor objective to exploitable routing gradients. Finally, richer multi-horizon critics, including hierarchical predictive-coding style world models, may preserve the representational benefit of distributed temporal prediction while keeping policy improvement aligned with the intended long-horizon target.

\bibliographystyle{unsrtnat}
\bibliography{references}

\end{document}